# A Utility-Theoretic Approach to Privacy in Online Services

**Andreas Krause**                                                          KRAUSEA@CALTECH.EDU
*California Institute of Technology,*
*1200 E California Blvd.,*
*Pasadena, CA 91125, USA*

**Eric Horvitz**                                                          HORVITZ@MICROSOFT.COM
*Microsoft Research,*
*One Microsoft Way,*
*Redmond, WA 98052-6399, USA*

## Abstract

Online offerings such as web search, news portals, and e-commerce applications face the challenge of providing high-quality service to a large, heterogeneous user base. Recent efforts have highlighted the potential to improve performance by introducing methods to personalize services based on special knowledge about users and their context. For example, a user's demographics, location, and past search and browsing may be useful in enhancing the results offered in response to web search queries. However, reasonable concerns about privacy by both users, providers, and government agencies acting on behalf of citizens, may limit access by services to such information. We introduce and explore an *economics of privacy* in personalization, where people can opt to share personal information, in a standing or on-demand manner, in return for expected enhancements in the quality of an online service. We focus on the example of web search and formulate realistic objective functions for search efficacy and privacy. We demonstrate how we can find a provably near-optimal optimization of the utility-privacy tradeoff in an efficient manner. We evaluate our methodology on data drawn from a log of the search activity of volunteer participants. We separately assess users preferences about privacy and utility via a large-scale survey, aimed at eliciting preferences about peoples willingness to trade the sharing of personal data in returns for gains in search efficiency. We show that a significant level of personalization can be achieved using a relatively small amount of information about users.

## 1. Introduction

Information about the preferences, activities, and demographic attributes of people using online applications can be leveraged to personalize the services for individuals and groups of users. For example, knowledge about the current locations of users performing web searches can help identify their informational goals. Researchers and organizations have pursued explicit and implicit methods for personalizing online services. For web search, explicit personalization methods rely on users indicating sets of topics of interest that are stored on a server or client. Implicit methods make use of information collected in the absence of user effort and awareness. Data collected implicitly in web search can include users locations and search activities, capturing such information as how people specify and reformulate queries and click, dwell, and navigate on results. Beyond web search, data collected about users in an implicit manner can be used to custom-tailor the behaviors of a broad spectrum of online applications from informational services like news summarizers to e-commerce services that provide access to online shopping, and that seek to maximize sales with targeted advertising.

The potential value of harnessing data about people to enhance online services coupled with the growing ubiquity of online services raises reasonable concerns about privacy. Both users and the hosts of online applications may benefit from the custom-tailoring of services. However, both may be uncomfortable with the access and use of personal information. There has been increasing discussion about incursions into the privacy of users implied by the general logging and storing of online data (Adar, 2007). Beyond general anxieties with sharing personal information, people may more specifically have concerns about becoming increasingly identifiable; as increasing amounts of personal data are acquired, users become members of increasingly smaller groups of people associated with the same attributes.

Most work to date on personalizing online services has either ignored the challenges of privacy and focused efforts solely on maximizing utility (c.f., Sugiyama, Hatano, & Ikoma, 2004) or has completely bypassed the use of personal data. One vein of research has explored the feasibility of personalizing services with methods that restrict the collection and analysis of personal data to users own computing devices (Horvitz, 2006). Research in this realm includes efforts to personalize web search by making use of the content stored on local machines, as captured within the index of a desktop search service (Teevan, Dumais, & Horvitz, 2005; Xu, Zhang, & Wang, 2007). Rather than cut off opportunities to make personal data available for enhancing online services or limit personalization to client-side analyses, we introduce and study utility-theoretic methods that balance the costs of sharing of personal data with online services in return for the benefits of personalization. Such a decision-theoretic perspective on privacy can allow systems to weigh the benefits of enhancements that come with adaptation with the costs of sensing and storage according to users preferences.

We characterize the utility of sharing attributes of private data via value-of-information analyses that take into consideration the preferences to users about the sharing of personal information. We explicitly quantify preferences about utility and privacy and then solve an optimization problem to find the best trade. Our approach is based on two fundamental observations. The first is that, for practical applications, the utility gained with sharing of personal data may often have a diminishing returns property; acquiring more information about a user adds decreasing amounts to the utility of personalization given what is already known about the users needs or intentions. On the contrary, the more information that is acquired about a user, the more concerning the breach of privacy becomes. For example, a set of individually non-identifying pieces of information may, when combined, hone down the user to membership in a small group, or even identify an individual. We map the properties of diminishing returns on utility and the concomitant accelerating costs of revelation to the combinatorial concepts of *submodularity* and *supermodularity*, respectively.

Although the economic perspective on privacy is relevant to a wide spectrum of applications, and to studies of the foundations of privacy more broadly, we shall illustrate the concepts in application to personalizing web search. We employ a probabilistic model to predict the web page that a searcher is going to visit given the search query and attributes describing the user. We define the utility of a set of personal attributes by the focusing power of the information gained with respect to the prediction task. Similarly, we use the same probabilistic model to quantify the risk of identifying users given a set of personal attributes. We then combine the utility and cost functions into a single objective function, which we use to find a small set of attributes which maximally increases the likelihood of predicting the target website, while making identification of the user as difficult as possible. The challenges of this optimization are in identifying the benefits and costs of sharing information and grappling with the computational hardness of the analysis. Solving for the best set of attributes for users to reveal (and hence for the optimal setting of the utility-privacy tradeoff) is an NP-hard search problem, and thus intractable in general for large sets of attributes. We shall demonstrate how we can use the submodularity of the utility and supermodularity of privacy in order to find a near-optimal tradeoff efficiently. To our knowledge, no existing approach (such as LeFevre, DeWitt, & Ramakrishnan, 2006; Chen, LeFevre, & Ramakrishnan, 2007; Hore & R. Jammalamadaka, 2007) provides such theoretical guarantees. We evaluate our approach on real-world search log data, as well as from data collected from a user study with over 1,400 participants focused on the elicitation of preferences about sharing sensitive information. Our results indicate the existence of prominent "sweet spots" in the utility-privacy tradeoff curve, at which most of the utility can be achieved with the sharing of a minimal amount of private information.

The manuscript is organized as follows. In Section 2, we formalize the utility-privacy tradeoff as an optimization problem and introduce objective functions. Section 3 identifies the submodular and supermodular structure of the utility and cost functions. In Section 4, we introduce an algorithm for finding a near-optimal solution, which exploits this combinatorial structure. In Section 6, we describe the experimental design of our user study. Section 5 describes the experimental setup, and Section 7 presents empirical evaluation of our approach on real-world search data. Section 8 presents related work. Section 9 reviews approaches to deploying the methodology described in this paper, and Section 10 presents a summary and conclusions.

## 2. Privacy-Aware Personalization

We consider the challenge of personalization as diagnosis under uncertainty: We seek to predict a searcher's information goals, given such noisy clues as query terms and potentially additional attributes that describe users and their interests and activities. We frame the challenge probabilistically (as done, e.g., in Dou, Song, & Wen, 2007; Downey, Dumais, & Horvitz, 2007 in the search context), by modeling a joint distribution $P$ over random variables, which comprise the target intention $X$, some request-specific attributes (e.g., the query term) $Q$, the identity of the user $Y$, and several attributes $\mathcal{V} = \{V_1, V_2, \ldots, V_m\}$ containing private information. Such attributes include user-specific variables (such as demographic information, search history, word frequencies on the local machine, etc.) and request-specific variables (such as the period of time since an identical query was submitted). We describe the concrete attributes used in this work for the web search context in Section 5. Additional examples are described by Downey et al. (2007) and Teevan et al. (2005). We shall describe the use of statistical techniques to learn a predictive model $P$ from training data for frequent queries. Then, we present methods for trading off utility and privacy in the context of the model.

### 2.1 Utility of Accessing Private Data

Upon receiving a new request $Q$, and given a subset $\mathcal{A} \subseteq \mathcal{V}$ of the attributes, we can use the probabilistic model to predict the target intention by performing inference, computing the conditional distribution $P(X \mid Q, \mathcal{A})$. Then, we use this distribution to inform the decision of, *e.g.*, which search results to present to the user. We use the notation $\#intents$ to refer to the domain size of $X$ (e.g., the maximum number of different webpages clicked on by the users). The hope in personalization is that additional knowledge about the user (i.e., the observed set of attributes $\mathcal{A}$) will help to simplify the prediction task, via reducing the uncertainty in $P(X \mid Q, \mathcal{A})$. Based on this intuition, we quantify the uncertainty in our prediction using the conditional Shannon entropy (c.f., Cover & Thomas, 1991) associated with the variance in target web sites following queries,

$$H(X \mid Q, \mathcal{A}) = -\sum_{x,q,\mathbf{a}} P(x, q, \mathbf{a}) \log_2 P(x \mid q, \mathbf{a}).$$

Hence, for any subset $\mathcal{A} \subseteq \mathcal{V}$, we define its utility $U(\mathcal{A})$ to be the *information gain*, i.e., expected entropy reduction achieved by observing $\mathcal{A}$:

$$\begin{aligned} U(\mathcal{A}) &= H(X \mid Q) - H(X \mid Q, \mathcal{A}) \\ &= -\sum_{x,q,\mathbf{a}} P(x, q, \mathbf{a}) \left[\log_2 P(x \mid q) - \log_2 P(x \mid q, \mathbf{a})\right]. \end{aligned}$$

Such *click entropy* has been previously been found effective by Dou et al. (2007).

## 2.2 Cost of Sharing Private Data

Several different models of privacy have been proposed in prior work (c.f., Sweeney, 2002; Machanava-jjhala, Kifer, Gehrke, & Venkitasubramaniam, 2006; Dwork, 2006). Our cost function is motivated by the consideration that sets of attributes $\mathcal{A} \subseteq \mathcal{V}$ should be preferred that make identification of individuals as difficult as possible. We can consider the observed attributes $\mathcal{A}$ as noisy observations of the (unobserved) identity $Y = y$ of the user. Intuitively, we want to associate high cost $C(\mathcal{A})$ with sets $\mathcal{A}$ which allow accurate prediction of $Y$ given $\mathcal{A}$, and low cost for sets $\mathcal{A}$ for which the conditional distributions $P(Y \mid \mathcal{A})$ are highly uncertain. For a distribution $P(Y)$ over users, we hence define an *identifiability loss* function $L(P(Y))$ which maps probability distributions over users $Y$ to the real numbers. $L$ is chosen in a way, such that if there exists a user $y$ such that $P(Y = y)$ is close to 1, then the loss $L(P(Y))$ is very large. If $P(Y)$ is the uniform distribution, then $L(P(Y))$ is close to 0. We will explore different loss functions $L$ below. Based on such loss functions, we define the *identifiability cost* $I(\mathcal{A})$ as the expected loss of the conditional distributions $P(Y \mid \mathcal{A} = \mathbf{a})$, where the expectation is taken over the observations $\mathcal{A} = \mathbf{a}$[1]:

$$I(\mathcal{A}) = \sum_{\mathbf{a}} P(\mathbf{a}) L(P(Y \mid \mathcal{A} = \mathbf{a})).$$

In addition to identifiability, we introduce an additional additive cost component $S(\mathcal{A}) = \sum_{a \in \mathcal{A}} s(a)$, where $s(a) \geq 0$ is a nonnegative quantity modeling the subjective *sensitivity* of attribute $a$, and other additive costs, such as data acquisition cost, etc. The final cost function $C(\mathcal{A})$ is a combination of the identifiability cost $I(\mathcal{A})$ and sensitivity $S(\mathcal{A})$, i.e., $C(\mathcal{A}) = I(\mathcal{A}) + S(\mathcal{A})$.

### 2.2.1 IDENTIFIABILITY LOSS FUNCTIONS

There are several ways to quantify identifiability. One approach to representing the loss L is with the negative entropy of the distribution $P(Y)$ of the user's identity $Y$[2], as used to quantify utility. However, in the context of privacy, this choice is rather poor: Consider a case where we seek to quantify the cost associated with the change in identifiability of the user that comes with learning the searchers gender. Assuming an equal distribution of males and females, learning the gender of the searcher would halve the space of possible searchers, hence increasing the entropy loss by one. However, this increase is independent of whether we start with (a) one billion or (b) only two searchers. In contrast to the influence on utility, where halving the search space of pages to consider is a very large gain, independent of the number of pages we start with (Dou et al., 2007), such a diminishment of anonymity is enormous: In case (a), an adversary trying to identify the searcher based on knowing their gender has almost no chance of success, whereas in case (b) they would always identify the person. Motivated by this consideration, we represent the privacy cost in our experiments as the *maxprob* loss (Chen et al., 2007), $L_m(P(Y)) = \max_y P(y)$. This loss function can be interpreted as follows: An adversary seeks to identify the user $Y$, and predicts the most likely user, and receives one unit reward if the user is guessed correctly, and 0 otherwise. The identifiability cost

$$I_m(\mathcal{A}) = \sum_{\mathbf{a}} P(\mathbf{a}) \max_y (P(y \mid \mathcal{A} = \mathbf{a}))$$

then is the expected win obtained by the adversary. This objective function makes most sense if we believe the adversary only has access to the same data sources as we do (and thus the probability distribution $P$ captures the assumptions about the adversary's inferences). We also consider the cost function

$$I_\ell(\mathcal{A}) = -\sum_{\mathbf{a}} P(\mathbf{a}) \log\Big(1 - \max_y(P(y \mid \mathcal{A} = \mathbf{a}))\Big).$$

---

1. Similarly, we can additionally take the expectation over the request $Q$
2. Note that instead of the user's identity (composed of all attribute values) in principle $Y$ could refer to a particular sensitive attribute (such as, e.g., sexual orientation).

$I_\ell$ is a "rescaled" variant of the *maxprob* loss, with the property that certainty (i.e., $\max_y(P(y \mid \mathcal{A} = \mathbf{a}) \approx 1$) is more severely penalized.

Another criterion for identifiability is *k-anonymity* (Sweeney, 2002). With this measure, a data set is called *k-anonymous*, if any combination of attributes is matched by at least $k$ people. We can define a probabilistic notion of k-anonymity, $I_k$, by using the loss function $L_k(P(Y))$ which is 1 if $P$ is nonzero for less than $k$ values of $Y$, 0 otherwise. The identifiability cost is then

$$I_k(\mathcal{A}) = \sum_{\mathbf{a}} P(\mathbf{a}) L_k(P(Y \mid \mathcal{A} = \mathbf{a})).$$

$I_k(\mathcal{A})$ can be interpreted as the expected number of violations of k-anonymity; a database (empirical distribution over users) is k-anonymous if and only if $I_k(\mathcal{A}) = 0$. See Lebanon, Scannapieco, Fouad, and Bertino (2009) for a justification of using a decision-theoretic analysis to quantify privacy and how inferential attacks through side information can be handled.

We experimentally compare the cost metrics in Section 7.1.

## 2.3 Optimizing the Utility-Privacy Tradeoff

Previously, we described how we can quantify the utility $U(\mathcal{A})$ for any given set of attributes $\mathcal{A}$, and its associated privacy cost $C(\mathcal{A})$. Our goal is to find a set $\mathcal{A}$, for which $U(\mathcal{A})$ is as large as possible, while keeping $C(\mathcal{A})$ as small as possible. To optimize utility for users under this tradeoff, we use scalarization (Boyd & Vandenberghe, 2004), and define a new, scalar objective

$$F_\lambda(\mathcal{A}) = U(\mathcal{A}) - \lambda C(\mathcal{A}).$$

Hereby, $\lambda$ plays the role of a privacy-to-utility conversion factor. The goal is to solve the following optimization problem:

$$\mathcal{A}_\lambda^* = \underset{\mathcal{A}}{\operatorname{argmax}} F_\lambda(\mathcal{A}) \tag{2.1}$$

By varying $\lambda$, we can find different solutions $\mathcal{A}_\lambda^*$. If we choose a very small $\lambda$, we find solutions with higher utility and higher cost; large values of $\lambda$ will lead to lower utility, but also lower privacy cost.

If the set of attributes $\mathcal{V}$ is large, then (2.1) is a difficult search problem, as the number of subsets $\mathcal{A}$ grows exponentially in the size of $\mathcal{V}$. It can be shown that the solution to this problem is hard even to approximate:

**Theorem 2.1.** *If there is a constant $\alpha > (1-1/e)$ and there exists an algorithm which is guaranteed to find a set $\mathcal{A}'$ such that $F_1(\mathcal{A}') \geq \alpha \max_{\mathcal{A}} F_1(\mathcal{A})$, then $P = NP$.*

The proofs of all theorems are presented in the Appendix. Given the complexity, we cannot expect to find a solution $\mathcal{A}^*$ efficiently which achieves even slightly more than $(1 - 1/e) \approx 63\%$ of the optimal score. However, we can find a solution which is guaranteed to achieve at least $1/3$ of the optimal value.

## 2.4 Hard Constraints on Privacy Cost

Instead of optimizing the tradeoff $F_\lambda(\mathcal{A}) = U(\mathcal{A}) - \lambda C(\mathcal{A})$, one may be interested in maximizing the utility $U(\mathcal{A})$ subject to a hard constraint on the cost $C(\mathcal{A})$, i.e., solve

$$\mathcal{A}_\lambda^* = \underset{\mathcal{A}}{\operatorname{argmax}} U(\mathcal{A}) \text{ s.t. } C(\mathcal{A}) \leq B, \tag{2.2}$$

for some value of $B \geq 0$. For example, users may be interested in maximizing utility while enforcing $k$-anonymity (in which case we would constrain $I_k(\mathcal{A}) \leq 0$). This solution would then provide per-user guarantees, i.e., $k$-anonymity is never violated. In principle, one could solve the tradeoff $F_\lambda$ (i.e., problem (2.1)) for different values of $\lambda$ and then, e.g., using binary search, choose $\lambda$ that maximizes $U(\mathcal{A})$ among all feasible solutions (i.e., $C(\mathcal{A}) \leq B$). In a sense, (2.1) can be seen as a Lagrangian relaxation of (2.2).

In the following, we will focus on the tradeoff problem (2.1). Using the procedure described above, our approach may be useful to solve the constrained problem (2.2) in practice (however our approximation guarantees will only hold for problem (2.1)).

## 3. Properties of the Utility-Privacy Tradeoff

As mentioned above, we would expect intuitively that the more information we already have about a user (i.e., the larger $|\mathcal{A}|$), the less the observation of a new, previously unobserved, attribute would help with enhancing a service. The combinatorial notion of *submodularity* formally captures this intuition. A set function $G : 2^\mathcal{V} \to \mathbb{R}$ mapping subsets $\mathcal{A} \subseteq \mathcal{V}$ into the real numbers is called *submodular* (Nemhauser, Wolsey, & Fisher, 1978), if for all $\mathcal{A} \subseteq \mathcal{B} \subseteq \mathcal{V}$, and $V' \in \mathcal{V} \backslash \mathcal{B}$, it holds that $G(\mathcal{A} \cup \{V'\}) - G(\mathcal{A}) \geq G(\mathcal{B} \cup \{V'\}) - G(\mathcal{B})$, i.e., adding $V'$ to a set $\mathcal{A}$ increases $G$ more than adding $V'$ to a superset $\mathcal{B}$ of $\mathcal{A}$. $G$ is called *nondecreasing*, if for all $\mathcal{A} \subseteq \mathcal{B} \subseteq \mathcal{V}$ it holds that $G(\mathcal{A}) \leq G(\mathcal{B})$.

A result of Krause and Guestrin (2005) shows that, under certain common conditional independence conditions, the reduction in click entropy is submodular and nondecreasing:

**Theorem 3.1** (Krause & Guestrin, 2005). *If the attributes $\mathcal{V}$ are conditionally independent given $X$, i.e., $P(V_1, \ldots, V_m \mid X) = \prod_i P(V_i \mid X)$ then $U(\mathcal{A})$ is submodular in $\mathcal{A}$.*

We discussed earlier how we expect the privacy cost to behave differently: Adding a new attribute would likely make a stronger incursion into personal privacy when we know a great deal about a user, and less if we know little. This "increasing costs" property corresponds with the combinatorial notion of *supermodularity*: A set function $G : 2^{\mathcal{V}} \to \mathbb{R}$ is called *supermodular* (Nemhauser et al., 1978), if for all $\mathcal{A} \subseteq \mathcal{B} \subseteq \mathcal{V}$, and $V' \in \mathcal{V} \setminus V$, it holds that $G(\mathcal{A} \cup \{V'\}) - G(\mathcal{A}) \leq G(\mathcal{B} \cup \{V'\}) - G(\mathcal{B})$, i.e., adding $V'$ to a large set $\mathcal{B}$ increases $G$ more than adding $V'$ to a subset $\mathcal{A}$ of $\mathcal{B}$. In fact, we can prove that the *maxprob* identifiability cost function introduced in Section 2 is supermodular.

**Theorem 3.2.** *Assume, the attributes $\mathcal{V}$ are marginally independent, and the user $Y$ is completely characterized by the attributes, i.e., $Y = (\mathcal{V})$. Then the* maxprob *loss $I_m(\mathcal{A})$ is supermodular in $\mathcal{A}$.*

Note that the attribute sensitivity $S(\mathcal{A})$ is per definition additive and hence supermodular as well. Thus, as a positive linear combination of supermodular functions, $C(\mathcal{A}) = I(\mathcal{A}) + S(\mathcal{A})$ is supermodular in $\mathcal{A}$ for $I(\mathcal{A}) = I_m(\mathcal{A})$. In our empirical evaluation, we verify the submodularity of $U(\mathcal{A})$ and supermodularity $C(\mathcal{A})$ even without the assumptions made by Theorem 3.1 and Theorem 3.2.

Motivated by the above insights about the combinatorial properties of utility and privacy, we formulate a general approach to trading off utility and privacy. We only assume that the utility $U(\mathcal{A})$ is a submodular set function, and that $C(\mathcal{A})$ is a supermodular set function. We define the general utility-privacy tradeoff problem as follows:

**Problem 3.3.** *Given a set $\mathcal{V}$ of possible attributes to select, a nondecreasing submodular utility function $U(\mathcal{A})$, a nondecreasing supermodular cost function $C(\mathcal{A})$, and a constant $\lambda \geq 0$, our goal is to find a set $\mathcal{A}^*$ such that*

$$\mathcal{A}^* = \operatorname*{argmax}_{\mathcal{A}} F_\lambda(\mathcal{A}) = \operatorname*{argmax}_{\mathcal{A}} U(\mathcal{A}) - \lambda C(\mathcal{A}) \tag{3.1}$$

Since $C(\mathcal{A})$ is supermodular if and only if $-C(\mathcal{A})$ is submodular, and since nonnegative linear combinations of submodular set functions are submodular as well, the scalarized objective $F_\lambda(\mathcal{A}) = U(\mathcal{A}) - \lambda C(\mathcal{A})$ is submodular as well. Hence, problem (3.1) requires the maximization of a submodular set function.

## 4. Optimization Algorithms

As the number of subsets $\mathcal{A} \subseteq \mathcal{V}$ grows exponentially with the size $n$ of $\mathcal{V}$, and the NP-hardness of Problem (2.1), we cannot expect to find the optimal solution $\mathcal{A}^*$ efficiently. Theorem 2.1 shows that it is NP-hard to even approximate the optimal solution to better than a constant factor of $(1-1/e)$. A fundamental result by Nemhauser et al. (1978) characterized the performance of the simple greedy algorithm, which starts with the empty set $\mathcal{A} = \emptyset$ and myopically adds the attribute which increases the score the most, i.e., $\mathcal{A} \leftarrow \mathcal{A} \cup \{\operatorname{argmax}_{V'} F(\mathcal{A} \cup \{V'\})\}$, until $k$ elements have been selected (where $k$ is a specified constant). It was shown that, if $F$ is nondecreasing, submodular and $F(\emptyset) = 0$, then the greedy solution $\mathcal{A}_G$ satisfies $F(\mathcal{A}_G) \geq (1-1/e) \max_{|\mathcal{A}|=k} F(\mathcal{A})$, i.e., the greedy solution achieves a value of at least a factor of $1-1/e$ of the optimal solution. Although this result would allow us to perform such tasks as selecting a near-optimal set of $k$ private attributes maximizing the utility $U(\mathcal{A})$ (which satisfies the conditions of the result from Nemhauser et al., 1978), it unfortunately does not apply in the more general case, where the objective $F_\lambda(\mathcal{A})$ is *not* nondecreasing.

The problem of maximizing such *non-monotone* submodular functions has been resolved by Feige, Mirrokni, and Vondrak (2007). A local search algorithm, named LS, was proved to guarantee a near-optimal solution $\mathcal{A}_{LS}$, if $F$ is a nonnegative[3] (but not necessarily nondecreasing) submodular function:

1. Let $V^* \leftarrow \mathrm{argmax}_{V' \in \mathcal{V}} F(\{V'\})$ and init. $\mathcal{A} \leftarrow \{V^*\}$

2. If there exists an element $V' \in \mathcal{V} \setminus \mathcal{A}$ such that $F(\mathcal{A} \cup \{V'\}) > (1 + \frac{\varepsilon}{n^2})F(\mathcal{A})$, then let $\mathcal{A} \leftarrow \mathcal{A} \cup \{V'\}$, and repeat step 2.

3. If there exists an element $V' \in \mathcal{A}$ such that $F(\mathcal{A} \setminus \{V'\}) > (1 + \frac{\varepsilon}{n^2})F(\mathcal{A})$, then let $\mathcal{A} \leftarrow \mathcal{A} \setminus \{V'\}$, and go back to step 2.

4. Return $\mathcal{A}_{LS} \leftarrow \mathrm{argmax}\{F(\mathcal{A}), F(\mathcal{V} \setminus \mathcal{A})\}$.

This algorithm works in an iterative manner to add or remove an element $V'$ in order to increase the score, until no further improvement can be achieved. Feige et al. (2007) prove the following Theorem:

**Theorem 4.1** (Feige et al., 2007). *If $F$ is a nonnegative submodular function, then, for the solution $\mathcal{A}_{LS}$ returned by algorithm LS, it holds that*

$$F(\mathcal{A}_{LS}) \geq \left(\frac{1}{3} - \frac{\varepsilon}{n}\right) \max_{\mathcal{A}} F(\mathcal{A}).$$

*LS uses at most $\mathcal{O}(\frac{1}{\varepsilon}n^3 \log n)$ function evaluations.*

Hence, LS returns a solution $\mathcal{A}_{LS}$ achieving at least 1/3 of the optimal score.

### 4.1 An Efficient Implementation

The description of Algorithm LS allows some freedom for implementing the search in steps 2 and 3. In our implementation, we select in a greedy manner the element $V'$ which most increases the objective function. Furthermore, to speed up computation, we use *lazy evaluation* to find the greedy elements (Robertazzi & Schwartz, 1989).

The algorithm LLS (c.f., Figure 1) performs a sequence of upwards and downwards passes, adding and removing elements which improve $F_\lambda$ by at least a factor of $(1 + \frac{\varepsilon}{n^2})$. The upwards pass is done in a lazy manner—the increments $\delta_V$ are *lazily* updated only when necessary. We found that the lazy computation can reduce the running time by an order of magnitude. The correctness of this lazy procedure follows directly from submodularity: Submodularity implies that the $\delta_V$ are monotonically nonincreasing as more and more elements are added to $\mathcal{A}$. The lazy updates can also be applied for the greedy downward pass. We discovered that Algorithm 1 usually terminates after one single downwards pass, without removing a single element.

---

3. If $F$ takes negative values, then it can be normalized by considering $F'(\mathcal{A}) = F(\mathcal{A}) - F(\mathcal{V})$, which however can impact the approximation guarantees.

---

**Input**: Submodular function $F_\lambda$
**Output**: Near-optimal selection $\mathcal{A}$ of personal attributes
**begin**
    $\mathcal{A} \leftarrow \emptyset$;
    **repeat**
        $change = false$;
        /* Lazy greedy upward pass:                        */
        **foreach** $V \in \mathcal{V}$ **do** $\delta_V \leftarrow F_\lambda(\mathcal{A} \cup \{V\}) - F_\lambda(\mathcal{A})$; $current_V \leftarrow true$;
        **repeat**
            $V' \leftarrow \text{argmax}_{V \in \mathcal{V} \setminus \mathcal{A}} \delta_V$;
            **if** $current_{V'} = true$ **then**
                **if** $\delta_{V'} > (1 + \frac{\varepsilon}{n^2})F_\lambda(\mathcal{A})$ **then**
                    $\mathcal{A} \leftarrow \mathcal{A} \cup \{V'\}$; **foreach** $V \in \mathcal{V}$ **do** $current_V \leftarrow false$
                **else**
                    **break**;
                **end**
            **else**
                $\delta_{V'} \leftarrow F_\lambda(\mathcal{A} \cup \{V'\}) - F_\lambda(\mathcal{A})$; $current_{V'} \leftarrow true$;
            **end**
        **until** $\delta_{V'} \leq (1 + \frac{\varepsilon}{n^2})F_\lambda(\mathcal{A})$ ;
        /* Lazy greedy downward pass:                  */
        **foreach** $V \in \mathcal{A}$ **do** $\delta_V \leftarrow F_\lambda(\mathcal{A} \setminus \{V\}) - F_\lambda(\mathcal{A})$; $current_V \leftarrow true$;
        **repeat**
            $V' \leftarrow \text{argmax}_{V \in \mathcal{A}} \delta_V$;
            **if** $current_{V'} = true$ **then**
                **if** $\delta_{V'} > (1 + \frac{\varepsilon}{n^2})F_\lambda(\mathcal{A})$ **then**
                    $\mathcal{A} \leftarrow \mathcal{A} \setminus \{V'\}$; $change = true$; **foreach** $V \in \mathcal{A}$ **do** $current_V \leftarrow false$
                **else**
                    **break**;
                **end**
            **else**
                $\delta_{V'} \leftarrow F_\lambda(\mathcal{A} \setminus \{V'\}) - F_\lambda(\mathcal{A})$; $current_{V'} \leftarrow true$;
            **end**
        **until** $\delta_{V'} \leq (1 + \frac{\varepsilon}{n^2})F_\lambda(\mathcal{A})$ ;
    **until** $change = false$ ;
    **return** $\text{argmax}\{F_\lambda(\mathcal{A}), F_\lambda(\mathcal{V} \setminus \mathcal{A})\}$
**end**

**Algorithm 1**: The *lazy local search* (LLS) algorithm.

## 4.2 Evaluating Utility and Cost

To run LLS, we need to be able to efficiently evaluate the utility $U(\mathcal{A})$ and cost $C(\mathcal{A})$. In principle, we can compute the objective functions from the empirical distribution of the training data, by explicitly evaluating the sums defining $U(\mathcal{A})$ and $C(\mathcal{A})$ (c.f., Section 2). However, this approach is very inefficient — $\Omega(N^2)$ where $N$ is the number of training examples. Instead, we can estimate $U(\mathcal{A})$ and $C(\mathcal{A})$ by sampling. Krause and Guestrin (2005) show how the Hoeffding inequality (Hoeffding, 1963) can be used in order to approximately compute conditional entropies. The Hoeffding inequality allows us to acquire bounds on the number of samples needed in order to determine the expectation $\mathbb{E}[\hat{H}]$ of a random variable $\hat{H}$, if $\hat{H}$ is bounded. In the case of click entropy reduction, we use the random variable

$$\hat{H} \mid [Q = q, \mathcal{A} = \mathbf{a}] = H(X) - H(X \mid q, \mathbf{a}).$$

$\hat{H}$ is a deterministic function modeling the click entropy reduction if request $Q = q$ and attributes $\mathcal{A} = \mathbf{a}$ are observed. Since $Q$ and $\mathcal{A}$ are random variables, $\hat{H}$ is random as well. Since $\hat{H}$ is bounded between 0 and $\log_2(\#intents)$, the Hoeffding inequality can be applied, and the following holds:

**Lemma 4.2** (Krause & Guestrin, 2005). *For any $\varepsilon > 0$ and $\delta > 0$, we need*

$$\left\lceil \frac{1}{2} \left( \frac{\log_2(\#intents)}{\varepsilon} \right)^2 \log \frac{1}{\delta} \right\rceil$$

*samples in order to estimate $U(\mathcal{A})$ to absolute error $\varepsilon$ with confidence at least $1 - \delta$.*

For the identifiability loss $I(\mathcal{A})$, we can proceed in a similar manner. Both the maximum probability and the $k$-anonymity loss are bounded between 0 and 1. Using a similar argument as in the proof of Lemma 4.2, we have the following result:

**Lemma 4.3.** *For any $\varepsilon > 0$ and $\delta > 0$, we need $\left\lceil \frac{1}{2\varepsilon^2} \log \frac{1}{\delta} \right\rceil$ samples in order to estimate $C(\mathcal{A})$ to absolute error $\varepsilon$ with confidence at least $1 - \delta$.*

We can generalize Theorem 4.1 to also hold in the case where utility and cost are estimated up to small constant error. The following theorem summarizes the analysis:

**Theorem 4.4.** *If $\lambda$ such that $F_\lambda(\mathcal{V}) \geq 0$, then* LLS*, using sampling to estimate $C(\mathcal{A})$ and $U(\mathcal{A})$, computes a solution $\mathcal{A}_{LLS}$ such that*

$$F_\lambda(\mathcal{A}_{LLS}) \geq \left( \frac{1}{3} - \frac{\varepsilon}{n} \right) \max_{\mathcal{A}} F_\lambda(\mathcal{A}) - n\varepsilon_S,$$

*with probability at least $1 - \delta$. The algorithm uses at most*

$$\mathcal{O}\left( \frac{1}{\varepsilon} n^3 \log n \left( \frac{\log_2(\#intents)}{\varepsilon_S} \right)^2 \log \frac{1}{\delta n^3} \right)$$

*samples.*

Hence, the algorithm LLS will efficiently find a solution $\mathcal{A}_{LLS}$ which achieves at least a constant fraction of 1/3 times the value of the optimal solution.

### 4.3 Computing Online Bounds

The bounds provided by Theorem 4.4 are *offline*, in the sense that they can be stated before running Algorithm 1. We can use the submodularity of $U(\mathcal{A})$ and supermodularity of $C(\mathcal{A})$ additionally for computing *online* bounds on the performance of *any* algorithm.

**Theorem 4.5.** *Let $\mathcal{A}' \subseteq \mathcal{V}$ be an arbitrary set of attributes. For each $V \in \mathcal{V} \setminus \mathcal{A}$ define $\eta_V = U(\mathcal{A}' \cup \{V\}) - U(\mathcal{A}') - \lambda C(\{V\})$. Let $\mathcal{B} = \{V : \eta_V > 0\}$. Then*

$$\max_{\mathcal{A}} F_\lambda(\mathcal{A}) \leq U_\lambda(\mathcal{A}') + \sum_{V \in \mathcal{B}} \eta_V.$$

It is again possible to extend these bounds to the case where the objective function $F_\lambda$ is evaluated with small absolute error via sampling.

### 4.4 Finding the Optimal Solution

Although LLS allows us to find a near-optimal solution in polynomial time, submodularity of $F_\lambda$ can also be exploited to find an optimal solution in a more informed way, allowing us to bypass an exhaustive search through all exponentially many subsets $\mathcal{A}$. Existing algorithms for optimizing submodular functions include branch and bound search, e.g., in the *data-correcting algorithm* by Goldengorin, Sierksma, Tijssen, and Tso (1999), as well as mixed-integer programming (Nemhauser & Wolsey, 1981). These approaches also do not require nonnegativity of $F_\lambda$. The mixed-integer programming approach by Nemhauser et.al. effectively uses bounds similar to those presented in Theorem 4.5.

## 5. Search Log Data and Attributes

We estimate the utility $U(\mathcal{A})$ and cost $C(\mathcal{A})$ of sets of private attributes from data. We use search log data, based on a total of 247,684 queries performed by 9,523 users from 14 months between December 2005 and January 2007. The search data was obtained from users who had volunteered to participate in a public Microsoft data sharing program centering on the use of information about their search activities to enhance search. The data was filtered to include only those queries which had been performed by at least 30 different users, resulting in a total of 914 different queries. For the utility $U(\mathcal{A})$, we compute the average reduction in click entropy (in bits) with respect to the per-query distribution of web pages chosen, as defined in Section 2. From the demographic information and the search logs, we compute 31 different user / query specific attributes. In selecting our attributes, we chose a very coarse discretization. No attribute is represented by more than three bits, and most attributes are binary. We consider the following attributes, which are summarized in Table 1.

### 5.1 Demographic Attributes

The first set of attributes contains demographic information, which were voluntarily provided separately as part of signing up for a set of online services. The attributes contain gender, age group, occupation and region, each very coarsely discretized into at most three bits. Gender was specified by 86% of users, age by 94%, occupation by 57%. 44% of the users specified their gender as male, 42% as female. 8% of users asserted that they are less than 18 years, 53% between 18 and 55, and 33% 55 and older. Most users were from the US and Canada (53%), followed by 7% from the European Union and 3% from Asia. Other locations were unspecified. 10% of the searchers specified that they are students.

## 5.2 Search Activity Attributes

The next set of attributes contains features extracted from search history data. For each query, we determine whether the same query has been performed before (AQRY; 70% of the queries are repeated), as well as whether the searcher has visited the same webpage (ACLK) before (53% of the clicks). We consider the sequence of websites visited following a query, and associate the hostname of the first website on which the surfer dwells for at least 30 seconds as the intended target. The attribute AFRQ describes whether the user performed at least one query each day. 47% of the searchers performed at least one search per day. We also log the top-level domain (ATLV) determined by reverse DNS lookup of the query IP address, and used only the domains .net, .com, .org and .edu. 83% of the queries were associated with one of these domains. We determine if a user ever performed queries from at least 2 different zip codes (AZIP; true 31%), cities (ACTY; true 31%) and countries (ACRY; true 2%), by performing reverse DNS lookup of the query IP addresses. For each query, we store whether the query was performed during working hours (AWHR; between 7 am and 6 pm) and during workdays (AWDY; Mon-Fri) or weekend (Sat, Sun), without accounting for holidays. Workdays account for 73% of the queries, but only 40% of the queries were done during working hours.

## 5.3 Topic Interests

We looked up all websites visited by the user during 2006 in the 16 element top-level category of the Open Directory Project directory (`www.dmoz.org`). For each category, we use a binary attribute indicating whether the user has ever visited a website in that category (acronyms for topics are indicated with prefix T). Topic classification was available for 96% of the queries.

## 6. Survey on Privacy Preferences

Although identifiability can be an important part of privacy, people may have different preferences about sharing individual attributes (Olson, Grudin, & Horvitz, 2005). We set out to assess preferences about cost and benefits of the sharing of different kinds of personal data. Related work has explored elicitation of private information (c.f., Huberman, Adar, & Fine, 2005; Wattal, Telang, Mukhopadhyay, & Boatwright, 2005; Hann, Hui, Lee, & Png, 2002). We are not familiar with a similar study for the context of web search. Our survey was designed specifically to probe preferences about revealing different attributes of private data in return for increases in the utility of a service (in this case, in terms of enhanced search efficiency). As previous studies by Olson et al. (2005) show, willingness to share information greatly depends on the type of information being shared, with whom the information is shared, and how the information is going to be used. In designing the survey, we tried to be as specific as possible, by specifying a low-risk situation, in which the "personal information would be shared and used only with respect to a single specified query, and discarded immediately thereafter." Our survey contained questions both on the sensitivity of individual attributes and on concerns about identifiability. The survey was distributed within Microsoft Corporation via an online survey tool. We motivated people to take the survey by providing participants with a lottery where they could win a media player via a random drawing. The survey was open to worldwide entries, and we received a total of 1,451 responses. Again, we use acronyms to refer to the personal attributes, as defined in Table 1.

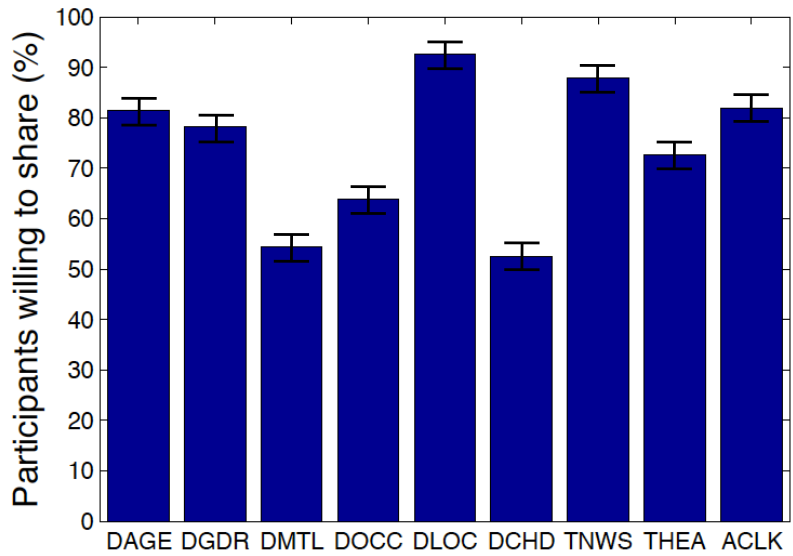

Figure 1: Willingness to share attributes

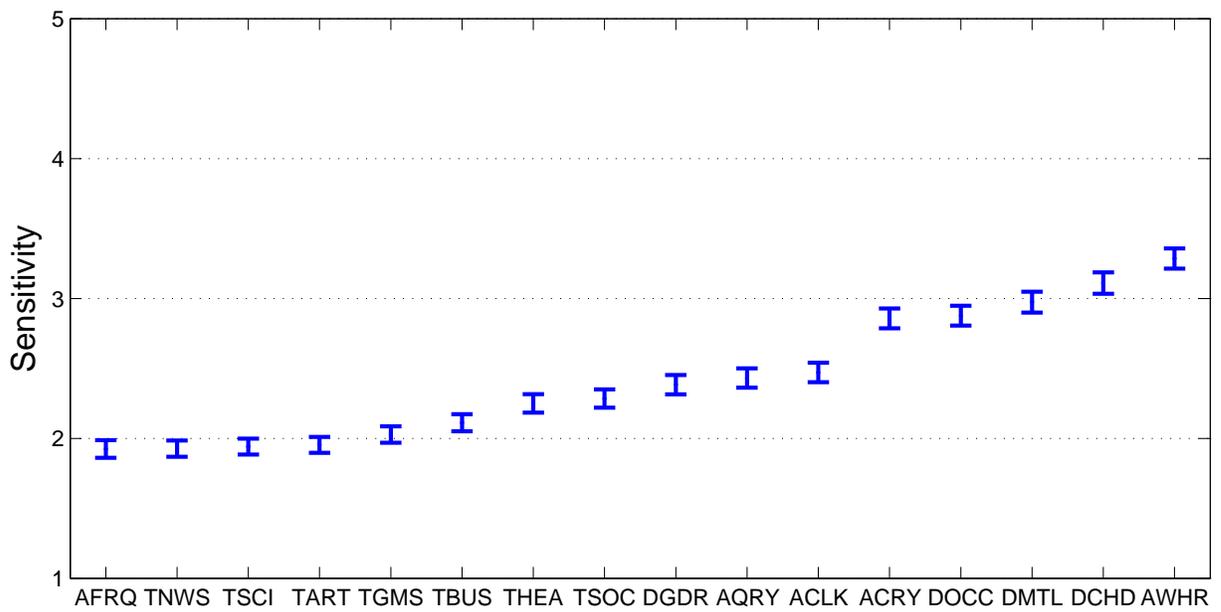

Figure 2: Sensitivity of individual attributes (with 95% confidence intervals)

## 6.1 Questions about Individual Attributes

First, we assessed attributes that participants would be willing to share with the search engine if revealing these attributes would "double search performance." Interestingly, for all attributes probed, more than 50% of study participants asserted that they would agree to share the information given the promised efficiency gain. Also, the sharing rates are very similar to those estimated in the survey (78% for gender, 82% for age, and 64% for occupation). The least willingness to share (54.4% and 52.6% respectively) was exhibited for marital status (DMTL) and whether the participant has children (DCHD), closely followed by occupation (63.9%). Most participants (92.7%) would rather share their region than share that their interests include news-related webpages (TNWS). Figure 1 presents the results for this question. We also asked the participants to classify the sensitivity of the attributes on a Likert scale from 1 (not very sensitive) to 5 (highly sensitive). The order of the questions was randomized. Figure 2 presents the results. The frequency of search engine usage (AFRQ) as well as very general topic interests, e.g., in news pages (TNWS), are considered to be of low sensitivity. Interestingly, we found that there are significant differences in preferences among participants even for sharing with a service interests in different topics; participants showed significantly greater sensitivity to sharing their interest in health or society related websites (THEA, TSOC) than in news or science-related pages (TNWS, TSCI). The biggest "jump" in sensitivity occurs between attributes ACLK, referring to sharing a repeated visit to same website, and ACRY, referring to having recently traveled internationally. We found that participants were most sensitive to sharing whether they are at work while performing a query (AWHR).

## 6.2 Questions about Identifiability

We also elicited preferences about sharing personal data at different degrees of precision and with different levels of identifiability. First, we sought to identify changes in the sensitivity associated with sharing personal data at increasingly higher resolution. More specifically, we inquired about the sensitivities of participants to sharing their ages at the level of groups of 20, 10, 5, and 1 years, and their exact birth dates. Similarly, we asked how sensitive participants would be to sharing their location at the region, country, state, city, zip code, or address levels of detail. Figures 3(a) and 3(b) present the mean sensitivity with 95% confidence intervals for this experiment. We also assessed participants sensitivities to having their search activity stored in different ways. More specifically, we asked users to assess their sensitivity about storing only a topic classification of the visited websites (i.e., whether a news- , business-, health-related etc. site was visited), storing all searches for 1 or 3 years, or storing all searches indefinitely. Lastly, we asked the participants how sensitive they would be, if, in spite of sharing the information, they would be guaranteed to remain indistinguishable from at least $k$ other people (thereby eliciting preferences about $k$ of $k$-anonymity). Here, we varied $k$ among 1, 10, 100, 1,000, 10,000, 100,000 and 1 million. Figure 3(c) presents the results of these experiments.

We draw a number of conclusions about the preferences of the population studied. First, storing search activity is generally considered more sensitive than sharing certain demographic information. For example, even storing only the topic classification of visited web pages is considered significantly more sensitive than sharing the city or birth year of a user. This result indicates that logging search activity is considered at least as threatening to privacy as sharing certain demographic information. The survey also shows that study participants have strong preferences about the granularity of the shared information. As explained below in Section 7.2, we can use the information obtained from this experiment to explicitly take into account peoples' preferences when trading off privacy and utility.

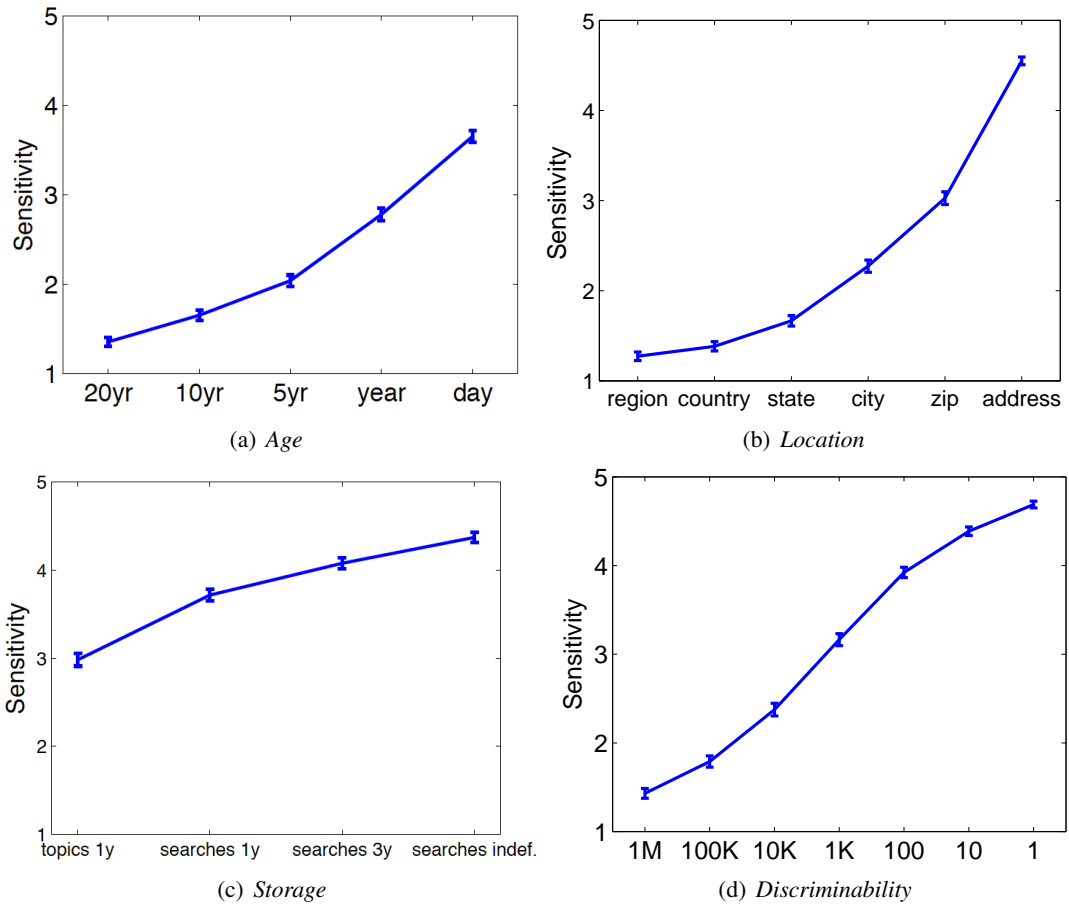

Figure 3: Sensitivity of sharing age (a) and location (b) under different levels of discretization. (c) Sensitivity of storing searches for varying amounts of time. (d) Sensitivity of $k$-discriminability levels (right). Plots show 95% confidence intervals.

### 6.3 Questions about Utility

In addition to assessing the sensitivity of sharing different kinds of personal information, we asked the participants to assess the degree of improvement they would require in order to share attributes of a given sensitivity level. More specifically, we asked: "How much would a search engine have to improve its performance, such that you would be willing to share information you consider 1/2/...". As response options, we offered average improvements by 25%, 50%, 100%, as well as the outcome of immediately presenting the desired page 95% of the time (which we associated with a speedup by a factor of 4). We also allowed participants to opt to never share information at a specified sensitivity level. These responses, in conjunction with the earlier sensitivity assessments, allowed us to establish sensitivity as a *common currency* of utility and cost. Figure 5 presents the median, 25 and 75-percentiles of the responses for the $k$-discriminability question and the above utility question.

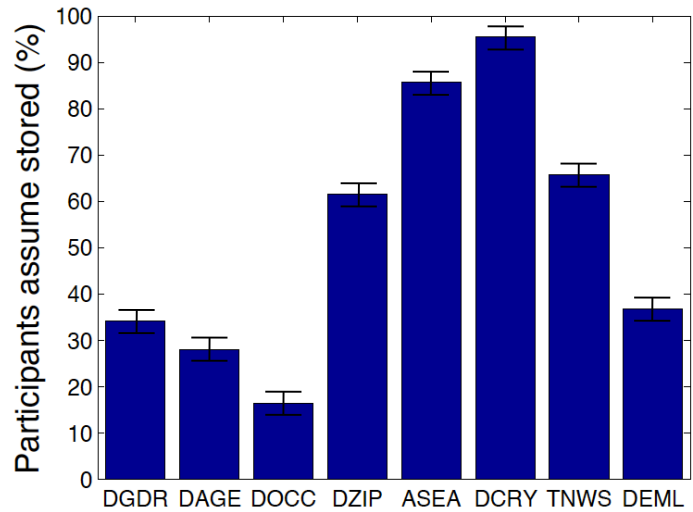

Figure 4: Which attributes are currently stored?

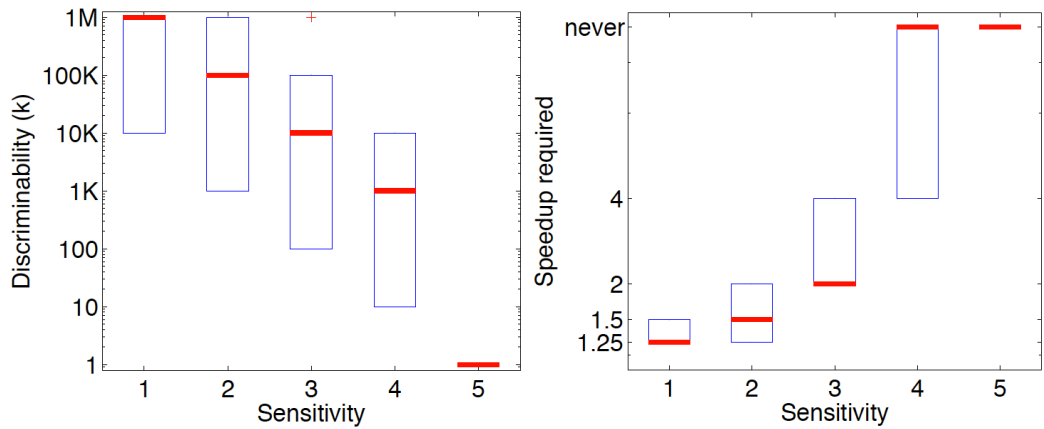

Figure 5: Using sensitivity as a common currency

### 6.4 Questions about the Current State

Additionally, we assessed the current understanding of the state of privacy in search. More specifically, we asked, whether the participants believe that most current search engines store the following attributes: DGDR, DAGE, DOCC, DZIP, ASEA, DCRY, TNWS, DEML. We found that most participants (96%) assume that search engines know the country (DCRY) a searcher is from. Also, most participants (86%) assume that all searches are stored by the search engines. Fewer participants believe that demographic information such as occupation (16%), age (28%) or gender (34%) are known to the search engine. This result is an important baseline in order to understand the sensitivity classification of the individual attributes.

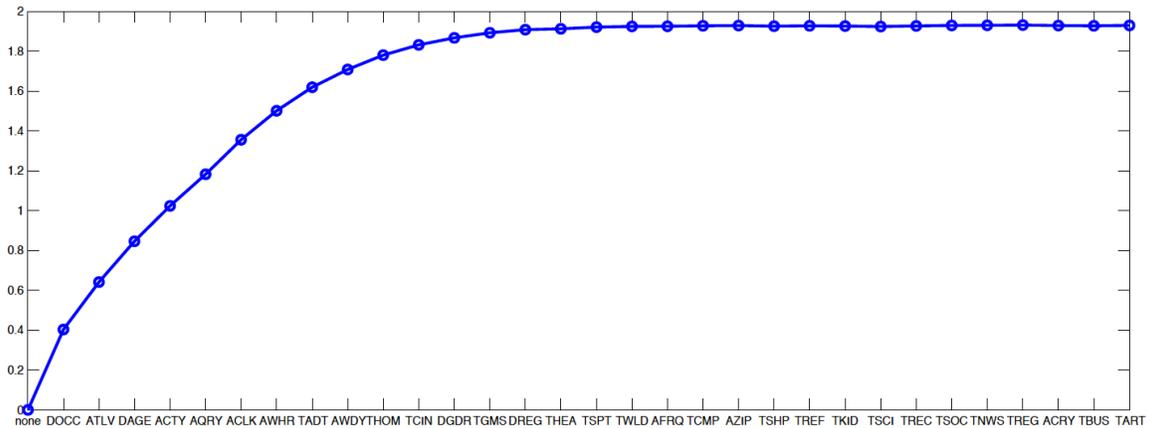

Figure 6: Utility (average click entropy reduction in bits) according to greedy ordering

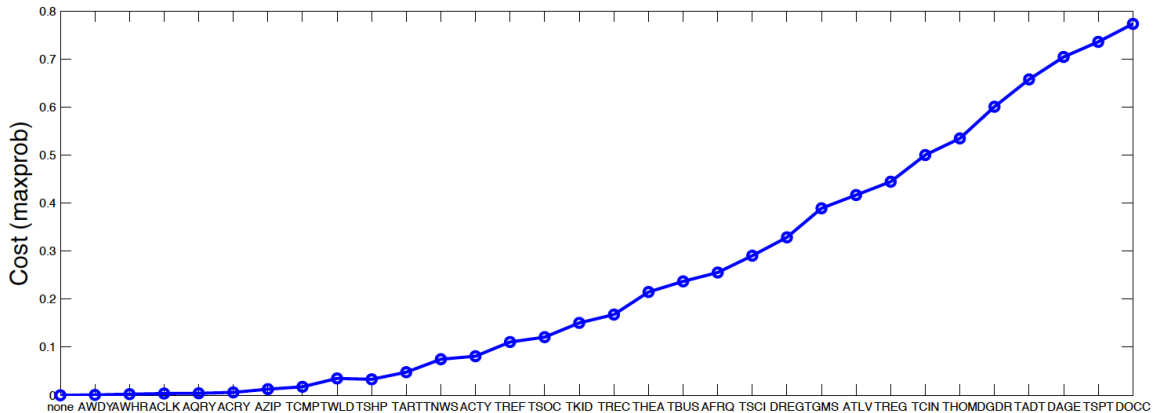

Figure 7: Cost according to greedy ordering

# 7. Results

We now describe our empirical results on calibrating and optimizing the utility-privacy tradeoff.

## 7.1 Computing Utility and Cost

We use the empirical distribution of the data described in Section 5, and evaluate utility and cost by sampling. Each sample is a row picked uniformly at random from the search logs. We then find all queries matching the selected attributes ($\mathcal{A} = \mathbf{a}$), and compute the conditional entropy of the click distribution, as well as the identifiability loss function. In order to avoid overfitting with sparse data, we applied Dirichlet smoothing. In our experiments, we use 1000 independent samples in order to estimate $U(\mathcal{A})$ and $I(\mathcal{A})$.

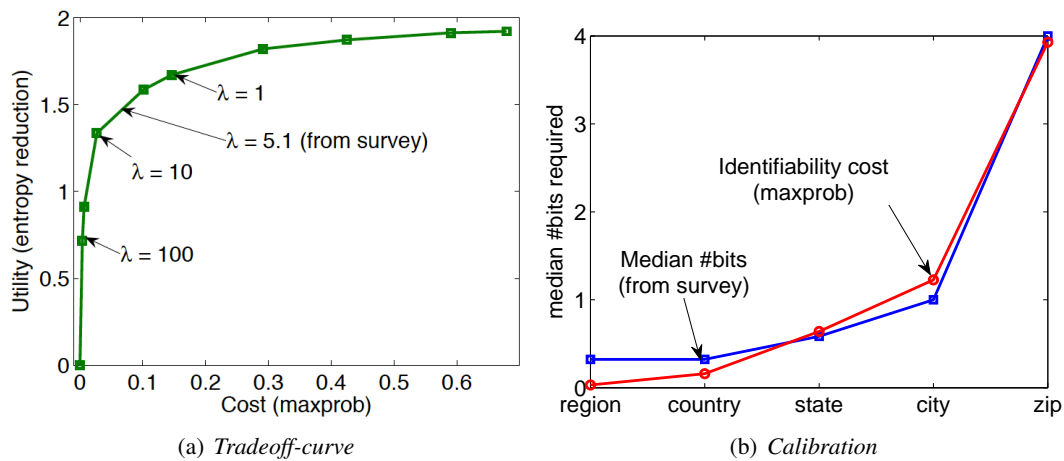

(a) *Tradeoff-curve*

(b) *Calibration*

Figure 8: (a) Tradeoff-curve for varying $\lambda$. (b) Calibrating the tradeoff.

We first apply the greedy algorithm to select an increasing number of attributes, maximizing the utility and ignoring the cost. Figure 6 presents the greedy ordering and the achieved entropy reductions. The greedy algorithm selects the attributes DOCC, ATLV, DAGE, ACTY, AQRY, ACLK, AWHR, TADT, AWDY, THOM, TCIN, DGDR, TGMS, TREG, in this order. After selecting these attributes, the utility does not increase significantly anymore. The entropy reduction levels off at roughly 1.92 bits. Figure 6 underscores the diminishing-returns property of click entropy reduction.

Similarly, we generate a greedy ordering of the attributes, in order of minimum incremental cost. Figure 7 presents the results of this experiment, using the maxprob cost metric. As expected, the curve looks convex (apart from small variations due to the sampling process). The cost initially increases very slowly, and the growth increases as more attributes are selected. This behavior empirically corroborates the supermodularity assumption for the cost metric.

Figure 9 compares the three cost metrics, as more and more attributes are selected. All three metrics initially behave qualitatively similarly. However, the $k$-anonymity metric flattens out after 25 out of 31 attributes have been selected. This is expected, as eventually enough personal information is available in order to (almost) always reduce the candidate set of people to less than $k = 100$. At this point, adding more attributes will not dramatically increase the cost anymore. However, when trading off utility and privacy, one is interested in solutions with small cost, and in this critical region, the cost function behaves supermodularly as well.

## 7.2 Calibrating the Tradeoff with Assessed Preferences

We now use the scalarization (3.1) to trade off utility and cost. For this optimization, we need to choose the tradeoff parameter $\lambda$. Instead of committing to a single value of $\lambda$, we generate solutions for increasing values of $\lambda$. For each such value, we use LLS to find an approximate solution, and plot its utility and cost. Figure 8(a) shows the tradeoff curve obtained from this experiment. We can see that this curve exhibits a prominent *knee*: For values $1 \le \lambda \le 10$, small increases of the utility lead of big increases in cost, and vice versa. Hence, at the knee, we can achieve near-maximal utility at near-minimum cost.

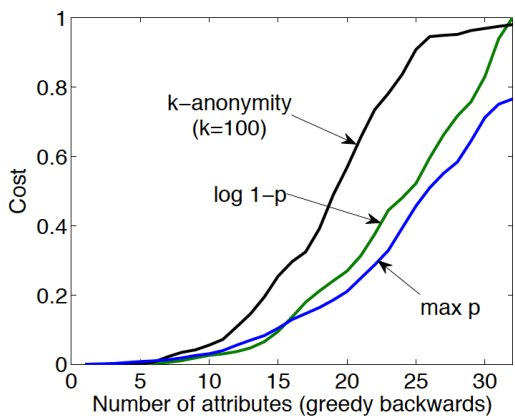

Figure 9: Cost comparison according to greedy ordering

To integrate peoples' preference in the analysis of the tradeoff, we perform the following *calibration* procedure. From the search log data, we determined how increasing the resolution of a person's location increases the privacy cost. We vary the location granularity from *region* (coarsest) to *zip code* (finest). For example, we compute the values $I_m(\{\text{zip code}\})$, $I_m(\{\text{city}\})$, etc. from data. As explained in Section 6.2, we had asked the subjects to assess the sensitivity of sharing their locations at different levels of precision. This approach also allows us put the identifiability cost $I(\mathcal{A})$ and sensitivity $S(\mathcal{A})$ into the same units.

Similarly, we assessed the amount of improvement in search performance that would be required in order to share attributes of a given sensitivity. We associate a number of bits with each level of improvement: A speedup by a factor of $x$ would require $\log_2 x$ bits (i.e., doubling the search performance would require 1 bit, etc.). We then concatenat the mappings from location granularity to sensitivity, and from sensitivity to utility (bits), and compute the median number of bits required for sharing each location granularity. Using this approach, we can put sensitivity $S(\mathcal{A})$ and utility $U(\mathcal{A})$ into the same units. Thereby, we effectively use sensitivity as a common currency between utility and cost. This procedure is (up to discretization) invariant of the particular scale (such as 1 to 5) used to assess sensitivity.

We perform a linear regression analysis to align the cost curve estimated from data with the curve obtained from the survey. The least-squares alignment is presented in Figure 8(b), and obtained for a value of $\lambda \approx 5.12$. Note that this value of $\lambda$ maps exactly into the sweet spot $1 \leq \lambda \leq 10$ of the tradeoff curve of Figure 8(b).

### 7.3 Optimizing the Utility-Privacy Tradeoff

Based on the calibration described above, our goal is to find a set of attributes $\mathcal{A}$ maximizing the calibrated objective $F_\lambda(\mathcal{A})$ according to (3.1).

First, we use the greedy algorithm to obtain an ordering of the attributes, similarly to the cases where we optimize utility and cost separately. Figure 10 presents the results of this experiment.

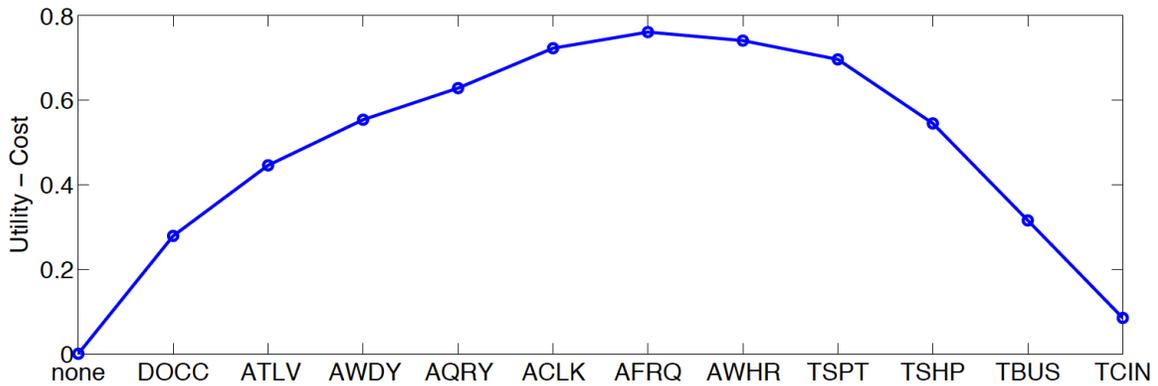

Figure 10: Greedy solutions for calibrated objective

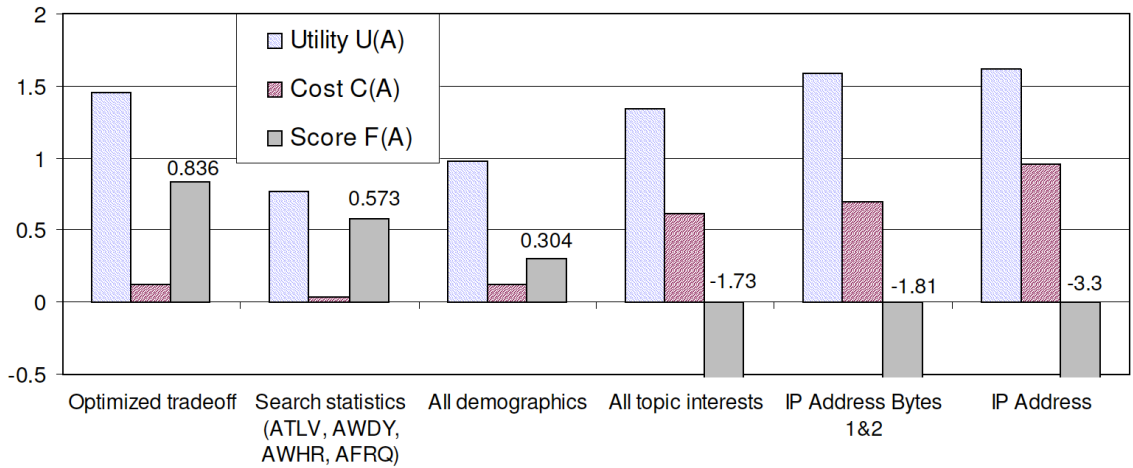

Figure 11: Comparison with heuristics

Instead of using the greedy algorithm, we can use LLS to approximately solve this optimization problem. The algorithm terminates after two upward and downward passes, with solution AFRQ, ATLV, AWDY, AQRY, ACLK, DAGE and TSPT. Note that the first element selected during the initial greedy upward pass is DOCC (the occupation), but it is discarded in the first downward pass again, since its individual sensitivity is quite high (c.f., Figure 2), and the additional information provided over the remaining 6 attributes is not high enough to warrant its presence in the optimal solution.

We also compared the optimized solution $\mathcal{A}_{opt}$ to the results of various heuristic procedures. For example, we compared it to the candidate solution $\mathcal{A}_{dem}$ where we select all demographic attributes (all attributes starting with D); $\mathcal{A}_{topic}$ where we select all topic interest attributes (starting with T); $\mathcal{A}_{search}$ including all search statistics (ATLV, AWDY, AWHR, AFRQ); $\mathcal{A}_{IP}$, the entire IP address or $\mathcal{A}_{IP2}$, the first two bytes of the IP address. Figure 11 presents the results of this comparison. The optimized solution $\mathcal{A}_{opt}$ obtains the best score of $0.83$, achieving a click entropy reduction of $\approx 1.4$.

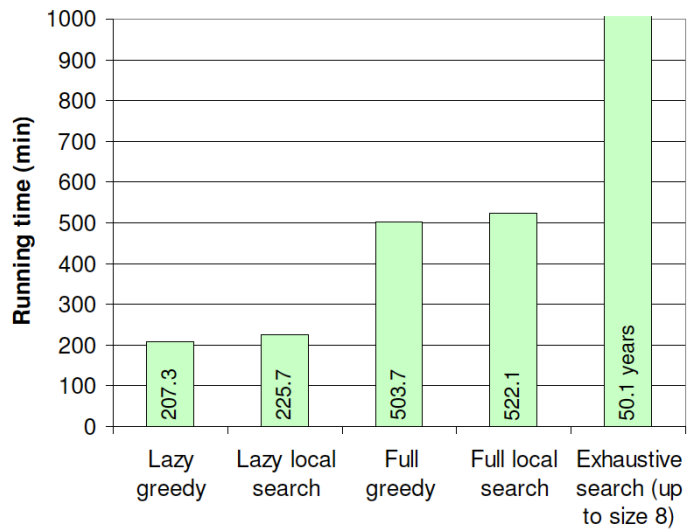

Figure 12: Running times

The search statistics $\mathcal{A}_{search}$ performs second best, with a score of 0.57, but achieving a drastically lower utility of only 0.8. Demographic information $\mathcal{A}_{dem}$ achieves higher utility of 0.95, but much higher cost, and hence an even lower total score of 0.3. Perhaps surprisingly, the collection of topic interests, $\mathcal{A}_{topic}$ results in negative total score of -1.73, achieving less utility than the optimized solution. We believe that the reason for this is that knowledge of the exact topic interest profile frequently suffices to uniquely identify a searcher. As expected, the IP address (even the first 2 bytes) are quite identifying in this data set, and hence has very high cost. This experiment shows that the optimization problem is non-trivial, and the optimized solution outperforms heuristic choices.

We also measured the running time of the algorithms, on a standard desktop PC (3 GHz) using a C#-implementation. Figure 12 presents the running times for the greedy and local search algorithms, both with and without the lazy evaluation trick. We note that the local search algorithm is not much slower than the greedy algorithm; frequently, only a single upward and downward pass are necessary, and only a small number of attributes have to be considered for elimination in the downward pass. We also note that using lazy evaluations drastically speeds up the running time—from 522.1 minutes without to 225.7 with lazy evaluations. To obtain a tradeoff curve like the one in Figure 8(b), the algorithm has to be run for each value of $\lambda$; hence any improvement is important. We also estimated the running time required for exhaustive search. Here, we considered only sets of size at most 8 (assuming that the optimal solution is of similar size as the approximate solution). Even this very optimistic estimate would require 50.1 years of computation time.

## 8. Related Work

This paper is an extended version of a paper that appeared at the 23rd Conference on Artificial Intelligence (AAAI) (Krause & Horvitz, 2008). The present version is significantly extended, including several additional experimental results, a detailed discussion of our LLS algorithm as well as new theoretical bounds (Section 4.3). We now review work related to the key concepts and methods presented in this paper.

| Label | Type | bits | Description |
|-------|------|------|-------------|
| DGDR | Demographic | 1 | Gender |
| DAGE | Demographic | 2 | Age group ($<18$, 18-50, $>50$) |
| DOCC | Demographic | 3 | Occupation (6 groups of related jobs) |
| DREG | Demographic | 2 | Region (4 geographic regions) |
| DMTL | Demographic | 1 | Marital status (*) |
| DCHD | Demographic | 1 | Whether the searcher has children or not (*) |
| AQRY | Activity | 1 | Performed same query before |
| ACLK | Activity | 1 | Visited same website before |
| AFRQ | Activity | 1 | User performs at least 1 query per day on average |
| AZIP | Activity | 1 | User performed queries from at least 2 different zip codes |
| ACTY | Activity | 1 | User performed queries from at least 2 different cities |
| ACRY | Activity | 1 | User performed queries from at least 2 different countries |
| AWHR | Activity | 1 | Current query performed during working hours |
| AWDY | Activity | 1 | Current query performed during workday / weekend |
| ATLV | Activity | 2 | Top-level domain of query IP address (.com, .net, .org, .edu) |
| TART | Topic | 1 | User previously visited arts related webpage |
| TADT | Topic | 1 | User previously visited webpage with adult content |
| TBUS | Topic | 1 | User previously visited business related webpage |
| TCMP | Topic | 1 | User previously visited compute related webpage |
| TGMS | Topic | 1 | User previously visited games related webpage |
| THEA | Topic | 1 | User previously visited health related webpage |
| THOM | Topic | 1 | User previously visited home related webpage |
| TKID | Topic | 1 | User previously visited kids / teens related webpage |
| TNWS | Topic | 1 | User previously visited news related webpage |
| TREC | Topic | 1 | User previously visited recreation related webpage |
| TREF | Topic | 1 | User previously visited reference related webpage |
| TREG | Topic | 1 | User previously visited webpage with regional content |
| TSCI | Topic | 1 | User previously visited science related webpage |
| TSHP | Topic | 1 | User previously visited shopping related webpage |
| TCIN | Topic | 1 | User previously visited consumer information webpage |
| TSOC | Topic | 1 | User previously visited society related webpage |
| TSPT | Topic | 1 | User previously visited sports related webpage |
| TWLD | Topic | 1 | User previously visited world related webpage |

Table 1: 33 Attributes used in our experiments. Attributes marked (*) were not available in search log data. Total number of bits = 38.

## 8.1 Personalized Search, Probabilistic Models for Search

The problem of personalized search has received a great deal of attention (c.f., Xu et al., 2007 for an overview of recent work). Teevan et al. considered the use of local term frequencies for re-ranking web search results (Teevan et al., 2005). The personal data is kept private as the analysis of personal information for re-ranking occurs entirely on the client. Recently, Xu et al. considered the problem of privacy preserving web search. Based on local information, a user profile is built, which is used for personalization. Two parameters called *minDetail* and *expRatio* are used to specify preferences about privacy. While similar in spirit, their approach does not tradeoff cost and benefit for sets of attributes as our approach does. Moreover, their approach does not consider the aspect of discriminability across users.

There have been efforts to employ probabilistic models in web search, e.g., for predicting relevance. Examples include the techniques proposed by Downey et al. (2007). The methods proposed in this paper can use any such probabilistic model, which allows to answer questions like "how much would my uncertainty decrease if I knew the searcher's gender".

## 8.2 Value of Information in Probabilistic Models

Optimizing value of information has been a cornerstone of principled approaches to information gathering (Howard, 1966; Lindley, 1956; Heckerman, Horvitz, & Middleton, 1993), and was popularized in decision analysis in the context of influence diagrams (Howard & Matheson, 1984). Several researchers (van der Gaag & Wessels, 1993; Cohn, Gharamani, & Jordan, 1996; Dittmer & Jensen, 1997; Kapoor, Horvitz, & Basu, 2007) suggested myopic, i.e., greedy approaches for selectively gathering observations. These algorithms typically do not have theoretical guarantees. Heckerman et al. (1993) propose a method to compute the maximum expected utility for specific sets of observations. They provide only large sample guarantees for the *evaluation* of a given sequence of observations, and use a heuristic without guarantees to select such sequences. Krause and Guestrin (2009) develop dynamic programming based algorithms for finding optimal sets of observations. However, their algorithms only apply to chain structured graphical models. Krause and Guestrin (2005) show that under certain conditional independence assumptions, the information gain is a submodular function, an observation that we build on in this paper. To our knowledge, this paper provides the first principled approach for efficiently, nonmyopically trading off value of information and privacy cost.

## 8.3 Valuation of Private Information

The problem of estimating sensitivity or monetary value of private information has been studied extensively, e.g., in the economics literature. Huberman et al. (2005) propose a second-price auction for estimating the sensitivity of demographic attributes. In a more application specific approach, Wattal et al. study, whether the inclusion of names or personal product preferences can enhance the effectiveness of email marketing (Wattal et al., 2005). Hann et al. (2002) study how economic incentives affect individual's preferences with different privacy policies, quantifying the value of disallowing the use of personal information. Kleinberg, Papadimitriou, and Raghavan (2001) quan-tify the value of private information using the Shapley value of a coalitional game. This approach provides an alternative, theoretically well-motivated way of eliciting the subjective cost of sharing a set of personal information, and could potentially be combined with our approach to optimize a utility-privacy tradeoff.

## 8.4 Mathematical Notions of Privacy

The field of mathematical (or cryptographic) privacy has been studied extensively (c.f., Adam & Wortmann, 1989 for a survey on early work). A pioneering result was the definition of $k$-anonymity, and the development of algorithms for maintaining this indiscriminability guarantee (Sweeney, 2002). Follow up work has led to other notions of indiscriminability, such as $l$-diversity (Machanavajjhala et al., 2006) etc. These notions describe properties of databases in isolation, and are sometimes called *non-interactive* (Dwork, 2006). Often, inference attacks using auxiliary knowledge are problematic in this context. Lebanon et al. (2009) consider decision theoretic variants of $k$-anonymity and related notions. Their approach allows to quantify the risk in (partially) releasing sanitized records. While they consider approaches to quantifying privacy cost, they do not present efficient algorithms with guarantees for trading off utility and privacy as we do in this paper.

Another important class of cryptographic approaches to privacy consider protection of privacy in the context of *interactive* analyses. In these approaches, a database contains an arbitrary amount of private information, which is securely guarded. Access to this database is enabled via a limited form of queries, which guarantee privacy. One prominent example of such techniques is the notion of *differential privacy* (Dwork, 2006). This notion quantifies the risk for each user incurred by participating in a database. Intuitively, queries are only allowed, if the corresponding results do not significantly depend on the presence or absence of individuals in the database. Blum et al. (2008) show how differential privacy (and a stronger variant called distributional privacy) can be achieved even in a non-interactive setting. However their approach is efficient only for a limited class of queries that, e.g., does not allow to estimate click probabilities as considered in this paper. Although these approaches provide crisp mathematical definitions of privacy, they are not designed with a specific notion of utility in mind. Rather, a specific privacy requirement is formulated, and analyses show the limits of data usage that are consistent with the requirement. We believe that the utility-theoretic approach to privacy is complementary to the cryptographic definitions of privacy. While our approach can be used to develop a privacy-aware design, existing techniques such as differential privacy can, e.g., be used to guard access to the private information. The utility-theoretic analyses also allow for new scenarios, such as identifying the value of specific private data for a specific context, and requesting this data in real time, for short-term usage. Such applications bypass the assumption of long-term, large-scale storage of private data that are explored within cryptographical analyses of privacy.

## 8.5 Utility-Privacy Tradeoff

Algorithmic approaches for optimizing the tradeoff have been considered in the area of privacy-preserving data publishing, which focuses on anonymization techniques. In these settings, a constraint on privacy is typically specified (in terms of $k$-anonymity, Sweeney, 2002, or $l$-diversity, Machanavajjhala et al., 2006) and then a recoding scheme is applied which maximizes a specified quality metric. Recent work involves the development of a greedy algorithm (LeFevre et al., 2006) and branch and bound search (Hore & R. Jammalamadaka, 2007). To our knowledge, these algorithms have no performance and approximation guarantees.

Our approach is different, in that we explicitly quantify both utility and cost in the same units, and optimize a single scalar value of information problem. Moreover, we use an application specific utility function, rather than a distortion metric. Furthermore, our algorithm is guaranteed to provide near-optimal solutions to the optimization problem.

## 9. Discussion

The proposed methodology could be implemented in a variety of ways. The most critical design choices are *where* and *when* the utility-privacy tradeoff is optimized. In the following, we shall explore these parameters in the context of personalized search.

### 9.1 Location of private information

Private information can be either located at the client (i.e., the recipient of the service, e.g., the web searcher), the server (i.e., the service provider, e.g., the search engine), or in the channel.

### 9.2 Client-side Private Information

In a client-side setting, private information is stored locally, and never shared with the network; this configuration has been considered by Teevan et al. (2005) and Xu et al. (2007). With such an approach, private information is used to re-rank search results locally, and never transmitted across the network, avoiding the risk of misuse or unintended release. However, the private information does not (easily) migrate with the user across multiple machines, and privacy concerns are effectively replaced by security concerns, requiring that the local machine not be compromised. The proposed methodology could be useful in mitigating security concerns; client services might be optimized to acquire and store only the most relevant information.

### 9.3 Server-side Private Information

On the server side, designs for privacy, especially with regard to the storage and usage of behavioral data, is of critical importance to services and their users.

The methods described can be used to balance user sensitivity and service utility, and to address such questions as which data should be logged, and which anonymization techniques should be used (c.f., Adar, 2007 for a discussion of the difficulties and possible options in anonymizing web search logs). Furthermore, specific personal (e.g., demographic) information could be voluntarily made available by the web searcher to the web service in form of, e.g., a user profile. In both settings, the proposed methodology could potentially be used to trade off privacy and utility.

### 9.4 Channel-side Private Information

We also foresee applications where private data is transmitted on a per-request basis. In this setting, the proposed methodology can be used to design which private "bits" to transmit to maximize utility while minimizing the privacy risk.

## 9.5 A Priori vs. Dynamic Tradeoff Optimization

Another important design parameter is *when* the tradeoff between privacy and utility is made.

### 9.5.1 A PRIORI TRADEOFF OPTIMIZATION / PRIVACY-AWARE SERVICE DESIGN

One option is to apply the proposed methodology *a priori*, i.e., before the system is being used. Here, the utility-privacy tradeoff would determine which information is being logged, collected in local profiles, or which private bits are made part of the protocol used to transmit service requests. This approach has the advantage that always the same private bits are used, and inference attacks combining private information from different requests are not possible.

### 9.5.2 DYNAMIC OPTIMIZATION

As we mentioned earlier, systems can be endowed with the ability to make dynamic recommendations to users or take actions in accordance with user preferences in the course of online activities. Dynamic variants of the methods we have presented can be used to identify the best subset of private information for sharing so as to optimize utility for a user. With web search, depending on the query, different kinds of personal information may be most helpful for disambiguation. The utility-theoretic approach to privacy could be used to determine needs in real time. Challenges with implementing such a real-time, interactive approach, include grappling with the possibility that inference attacks can combine the bits from different requests. Anonymization procedures can be designed to resist such challenges. Another challenge is computational efficiency, as solving an optimization problem may be required for each service request. The main advantage of employing a dynamic approach is that the average number of private bits (and hence the identifiability cost) per request could be lower than in the a priori privacy-aware design. The a priori design requires a collection of private bits which help disambiguating all queries versus context-specific situations. This interactive approach would additionally allow users to share private information on a per-request basis. Such interactive decisions also require considerations of how preferences about tradeoffs and private information are acquired from users—and will rely on the design of user interaction models and interfaces to provide users with awareness about opportunities for enhancing the value of services and controls for guiding the revelation of private information. Recent developments in *adaptive submodularity* may provide the methods necessary to achieve such dynamic, adaptive optimization of tradeoffs (Golovin & Krause, 2010).

## 9.6 Other Applications

The increasing cost property which characterizes the identifiability cost potentially arises in other contexts as well. For example, for many algorithms, the computational cost scales superlinearly in the number of dimensions / attributes considered. In such cases, the problem of trading off information (such as the click entropy reduction) and computation cost have the same combinatorial structure as the utility-privacy tradeoff addressed in this paper. Some of the algorithmic/computational considerations presented in this paper, such as the lazy local search algorithm, or sample complexity analyses, apply to such contexts as well.

## 10. Conclusions

We presented an approach for explicitly optimizing the utility-privacy tradeoff in personalized services such as web search. We showed that utility functions like click entropy reduction satisfy submodularity, an intuitive diminishing-returns property. In contrast, privacy concerns show supermodularity; the more private information we accrue, the faster sensitivity and the risk of identifiability grow. Based on the submodular utility and supermodular cost functions, we demonstrated how we can efficiently find a provably near-optimal utility-privacy tradeoff. We evaluated our methodology on real-world web search data. We demonstrated how the quantitative tradeoff can be calibrated according to personal preferences, obtained from user study with over 1,400 participants. Overall, we found that significant personalization can be achieved using only a small amount of information about users. We believe that the principles and methods employed in the utility-theoretic analysis of tradeoffs for web search have applicability to the personalization of a broad variety of online services. The results underscore the value of taking a decision-theoretic approach to privacy, where we seek to jointly understand the utility of personalization that can be achieved via access to information about users, and the preferences of users about the costs and benefits of selectively sharing their personal data with online services.

### Acknowledgments

Andreas Krause was an intern at Microsoft Research while this work was performed. We would like to thank the searchers who provided usage data for use in this research, the participants in our survey about privacy preferences, and the anonymous referees for their helpful comments and suggestions.

## Appendix A. Proofs

*Proof of Theorem 2.1.* Let $n = |cV|$. We prove the result for a choice of $I(\mathcal{A})$, where $I(\mathcal{A}) = 0$ if $|\mathcal{A}| < n/2$ and $I(\mathcal{A}) = 2 * M * (|\mathcal{A}| - n/2)$ otherwise, where $M = U(\mathcal{V})$ is the maximum achievable utility. In this case, $F_1(\mathcal{A}) = U(\mathcal{A}) \geq 0$ if $|\mathcal{A}| \leq n/2$, and $F_1(\mathcal{A}) < 0$ if $|\mathcal{A}| > n/2$. Hence,

$$\operatorname*{argmax}_{\mathcal{A}} F_1(\mathcal{A}) = \operatorname*{argmax}_{\mathcal{A}:|\mathcal{A}|\leq n/2} U(\mathcal{A}).$$

However, Krause and Guestrin (2005) show that if there were a polynomial time algorithm which is guaranteed to find a set $\mathcal{A}'$ such that $U(\mathcal{A}') \geq (1 - 1/e) \max_{\mathcal{A}:|\mathcal{A}|\leq n/2} U(\mathcal{A})$, then $P = NP$. $\qquad\square$

*Proof of Theorem 3.2.* Let $Y = (\mathcal{A}, \mathcal{B})$ (i.e., $\mathcal{V}$ is partitioned into $\mathcal{A}$ and $\mathcal{B}$). Then,

$$I_m(\mathcal{A}) = \sum_{\mathbf{a}} P(\mathbf{a}) \max_{\mathbf{b}} P(\mathbf{b} \mid \mathbf{a}) = \sum_{\mathbf{a}} P(\mathbf{a}) \max_{\mathbf{b}} P(\mathbf{b})$$

$$= \max_{\mathbf{b}} P(\mathbf{b}) = \max_{\mathbf{b}} \prod_i P(b_i) = \prod_i \max_{b_i} P(b_i) = \prod_i w_i$$

where $w_i = \max_{b_i} P(b_i)$. Now, $I_m(\mathcal{A})$ is nondecreasing in $\mathcal{A}$. Furthermore, for subsets $\mathcal{A}, \mathcal{A}' \subseteq \mathcal{V}$ and $\mathcal{B} = \mathcal{A} \cup \mathcal{A}'$, it holds that $I_m(\mathcal{B}) = \alpha I_m(\mathcal{A})$ for $\alpha \geq 1$. Furthermore, if $V \in \mathcal{V} \setminus \mathcal{B}$ and $w = \max_v P(V = v)$, then $I_m(\mathcal{B} \cup \{V\}) = w^{-1} I_m(\mathcal{B})$ and $I_m(\mathcal{A} \cup \{V\}) = w^{-1} I_m(\mathcal{A})$. Hence,

$$\frac{I_m(\mathcal{B} \cup \{V\}) - I_m(\mathcal{B})}{I_m(\mathcal{A} \cup \{V\}) - I_m(\mathcal{A})} = \frac{\alpha(w^{-1} - 1)}{w^{-1} - 1} = \alpha \geq 1,$$

which proves the supermodularity of $I_m(\mathcal{A})$. □

*Proof of Theorem 4.4.* Additive error carries through Lemma 3.3. and subsequently through the proof of Theorem 3.4. of Feige et al. (2007). In order to guarantee the confidence $1 - \delta$, we apply the union bound. The sample complexity then follows from Lemma 4.2 and Lemma 4.3. □

*Proof of Theorem 4.5.* Let $\mathcal{C}$ be an optimal solution. Since $U$ is nondecreasing,

$$
\begin{aligned}
F(\mathcal{C}) = U(\mathcal{C}) - \lambda C(\mathcal{C}) &\leq U(\mathcal{A}' \cup \mathcal{C}) - \lambda C(\mathcal{C}) \\
&\leq U(\mathcal{A}) + \sum_{V \in \mathcal{C}} [U(\mathcal{A} \cup \{V\}) - U(\mathcal{A})] - \lambda C(\mathcal{C}) \\
&\leq U(\mathcal{A}) + \sum_{V \in \mathcal{C}} [U(\mathcal{A} \cup \{V\}) - U(\mathcal{A}) - \lambda C(\{V\})] \\
&= U(\mathcal{A}') + \sum_{V \in \mathcal{C}} \eta_V \leq U(\mathcal{A}') + \sum_{V \in \mathcal{B}} \eta_V.
\end{aligned}
$$

□